\documentclass[preprint,12pt]{elsarticle}

\usepackage{amssymb}
\usepackage{amsmath}

\usepackage{booktabs}
\usepackage{caption}
\usepackage{subcaption}
\usepackage{multirow}

\usepackage{floatrow}
\usepackage{hyperref}
\usepackage{url}
\usepackage{graphicx}
\usepackage{makecell}

\usepackage{xcolor, array, colortbl}

\definecolor{deepred}{rgb}{0.698,0.133,0.133}
\definecolor{blue}{rgb}{0,0,1}

\definecolor{orange}{rgb}{1,0.38,0}
\definecolor{beige}{rgb}{0.639,0.58,0.502}

\definecolor{lightgray}{rgb}{.91,.91,.91}

\usepackage{graphicx}

\newcommand{\arr}[1][3pt]{\mathrel{%
   \vcenter{\hbox{\rule[-.2pt]{#1}{.4pt}}}%
   \mkern-5mu\hbox{\usefont{U}{lasy}{m}{n}\symbol{41}}}}

\journal{Neural Networks}

\begin{document}

\begin{frontmatter}

\title{Information-Theoretic Complementary Prompts for Improved Continual Text Classification}

\author[mymainaddress]{Duzhen Zhang\corref{co-first}}
\cortext[co-first]{Duzhen Zhang and Yong Ren are co-first authors of this paper. This work was done when Yong Ren was interning at Tencent, AI Lab, Beijing, China.}
\ead{duzhen.zhang@mbzuai.ac.ae}

\author[mythirdaddress]{Yong Ren\corref{co-first}}

\author[mythirdaddress]{Chenxing Li}

\author[myfifthaddress]{Dong Yu}

\author[myfourthaddress]{Tielin Zhang\corref{mycorrespondingauthor}}
\cortext[mycorrespondingauthor]{Corresponding author.}
\ead{zhangtielin@ion.ac.cn}

\address[mymainaddress]{Mohamed bin Zayed University of Artificial Intelligence, Abu Dhabi, UAE}
\address[myfourthaddress]{Center for Excellence in Brain Science and Intelligence Technology, Chinese Academy of Sciences, Shanghai, China}
\address[mythirdaddress]{Tencent AI Lab, Beijing, China}
\address[myfifthaddress]{Tencent AI Lab, Bellevue, USA}

\begin{abstract}

Continual Text Classification (CTC) aims to continuously classify new text data over time while minimizing catastrophic forgetting of previously acquired knowledge. However, existing methods often focus on task-specific knowledge, overlooking the importance of shared, task-agnostic knowledge.
Inspired by the complementary learning systems theory, which posits that humans learn continually through the interaction of two systems—the hippocampus, responsible for forming distinct representations of specific experiences, and the neocortex, which extracts more general and transferable representations from past experiences—we introduce Information-Theoretic Complementary Prompts (InfoComp), a novel approach for CTC.
InfoComp explicitly learns two distinct prompt spaces: P(rivate)-Prompt and S(hared)-Prompt. These respectively encode task-specific and task-invariant knowledge, enabling models to sequentially learn classification tasks without relying on data replay. To promote more informative prompt learning, InfoComp uses an information-theoretic framework that maximizes mutual information between different parameters (or encoded representations).
Within this framework, we design two novel loss functions: (1) to strengthen the accumulation of task-specific knowledge in P-Prompt, effectively mitigating catastrophic forgetting, and (2) to enhance the retention of task-invariant knowledge in S-Prompt, improving forward knowledge transfer. 
Extensive experiments on diverse CTC benchmarks show that our approach outperforms previous state-of-the-art methods.

\end{abstract}


\end{frontmatter}

\section{Introduction}

Mastering a wide range of tasks, accumulating experience, and preventing forgetfulness are fundamental characteristics of human-level intelligence. Despite the substantial advancements in Pretrained Language Models (PLMs) like BERT~\cite{devlin2019bert}, their performance still deteriorates when confronted with a sequence of downstream text classification tasks—a scenario known as Continual Text Classification (CTC)~\cite{de2019episodic,huang2021continual}. CTC presents two key challenges: (1) Catastrophic Forgetting (CF), where models tend to lose previously acquired knowledge when learning new tasks~\cite{robins1995catastrophic,mccloskey1989catastrophic,goodfellow2013empirical,zhai2024investigating,zhang2024balancing}, and (2) Forward Knowledge Transfer (FKT), which involves utilizing knowledge from earlier tasks to enhance the learning efficiency of subsequent tasks.

As the parameter scale of modern PLMs continues to expand, fine-tuning the entire model becomes increasingly impractical, which has led to growing interest in Parameter-Efficient Fine-Tuning (PEFT) techniques. Among these, Prompt Tuning (PT)~\cite{liu2022p} has emerged as a prominent solution. PT works by learning a soft prompt and appending it to the input of PLMs, while keeping the model itself frozen~\cite{liu2022p}. Previous research demonstrates that adapting PLMs to individual downstream tasks through prompt learning can achieve performance on par with full model fine-tuning, while using less than 0.01\% of the parameters~\cite{lester2021power,liu2023pre}.
Essentially, PT shifts the focus from modifying the model’s parameters to crafting prompts that guide the model’s learning for specific tasks. These prompts encapsulate task-specific knowledge, making PT more effective in utilizing frozen PLMs compared to traditional fine-tuning approaches. In the context of CTC, harnessing PT to acquire and retain knowledge holds great potential~\cite{wang2022learning}.

Recently, Progressive Prompt (ProgPrompt) has been introduced as an extension of PT for handling sequential downstream tasks, achieving State-Of-The-Art (SOTA) performance on various CTC benchmarks~\cite{razdaibiedina2022progressive}. ProgPrompt learns distinct prompts for each incoming task and reuses them during inference, effectively mitigating CF. Moreover, it facilitates FKT by incorporating previously learned prompts (which remain frozen) as inputs when learning a new task. 
However, ProgPrompt faces several significant limitations. First, it focuses exclusively on task-specific knowledge without leveraging the knowledge shared across tasks. Second, by directly concatenating all previously learned prompts for FKT, it inevitably introduces redundant information. Lastly, ProgPrompt maintains a continuously growing prompt list. As the number of tasks ($N$) increases, the list grows at a rate of $\mathcal{O}(N)$, while transformer-based PLMs have a computational complexity of $\mathcal{O}(N^2)$~\cite{vaswani2017attention}. This results in escalating costs for both training and inference as the number of tasks increases.

To overcome these limitations, we propose Information-Theoretic Complementary Prompts (InfoComp), a method designed to effectively guide PLMs in learning sequential downstream text classification tasks. Drawing inspiration from complementary learning systems~\cite{mcclelland1995there,kumaran2016learning}, we model our approach after the way humans engage in continual learning through the collaboration between the hippocampus and neocortex. The hippocampus excels at acquiring pattern-separated representations from specific experiences, while the neocortex captures more general and transferable representations from a sequence of past experiences~\cite{arani2021learning,pham2021dualnet,wang2022dualprompt}.
Based on this analogy, InfoComp explicitly learns two distinct prompt spaces: P(rivate)-Prompt and S(hared)-Prompt. The P-Prompt encodes task-specific knowledge, while the S-Prompt captures task-invariant knowledge shared across all tasks. Instead of concatenating all previously learned P-Prompts, InfoComp utilizes the task-invariant knowledge accumulated in the S-Prompt for FKT, thereby avoiding redundant information. When learning a new task, only the corresponding P-Prompt and the shared S-Prompt are required, ensuring that the prompt length remains constant as the number of tasks ($N$) increases.

To further improve the generation of more informative prompts, InfoComp frames soft PT as a process of maximizing Mutual Information (MI) between prompts and other model parameters, as well as between different encoded representations. Within this framework, we introduce two novel loss functions based on MI: (1) Enhancing task-specific knowledge in P-Prompt: We maximize the MI between the P-Prompt and the task-specific classifier. Since the classifier typically encapsulates critical information from the downstream task~\cite{wu2023infoprompt}, optimizing this MI allows the P-Prompt to capture richer task-relevant information, thereby more effectively mitigating CF. (2) Preserving task-invariant knowledge in S-Prompt: We maximize the MI between the encoded representations of the same input under the current and previous S-Prompt conditions. By doing so, we ensure better preservation of task-invariant knowledge, facilitating more effective FKT and promoting continuity of shared knowledge across sequential tasks.

Our contributions can be summarized as follows:
\begin{itemize}
\item We introduce InfoComp for CTC, which integrates P-Prompt and S-Prompt to effectively capture task-specific and task-invariant knowledge, respectively. This method is easy to implement and eliminates the need for data replay, making it particularly well-suited for real-world continual learning applications.

\item We advance prompt learning from an information-theoretic perspective by proposing two novel loss functions: one to enhance the acquisition of task-specific knowledge in P-Prompt, and the other to preserve task-invariant knowledge in S-Prompt.

\item  Extensive experiments on diverse CTC benchmarks, including both standard setups and more challenging scenarios with longer task sequences, demonstrate that InfoComp significantly outperforms previous SOTA methods.

\end{itemize}

\section{Related Work}

\subsection{Continual Learning}

Continual Learning (CL) aims to develop algorithms capable of progressively accumulating knowledge from a sequence of tasks~\cite{DBLP:journals/corr/abs-2302-00487,zheng2025lifelong,dong2023federated,zhang2025federated,dong2024continually}. Traditional CL methods are generally divided into three categories: replay-based, regularization-based, and architecture-based approaches.
Replay-based methods utilize a rehearsal buffer to retrain on a subset of previous examples~\cite{rebuffi2017icarl,chaudhry2018efficient}. However, their effectiveness diminishes as the buffer size decreases~\cite{DBLP:conf/iccv/ChaLS21}, and they are unsuitable for scenarios where data privacy is a concern~\cite{DBLP:conf/ccs/ShokriS15}. 
Regularization-based methods address forgetting by imposing constraints on network weights~\cite{kirkpatrick2017overcoming,farajtabar2020orthogonal,huang2021continual}, intermediate features~\cite{hou2019learning,zhang2023continual}, or output probabilities~\cite{li2017learning,zhang2023task}. While these approaches mitigate forgetting, they often incur additional storage costs for gradients or models from previous tasks, and they tend to struggle with long task sequences.
Architecture-based methods, on the other hand, dynamically allocate isolated modules to store knowledge from different tasks, thereby minimizing interference~\cite{rusu2016progressive,yoon2018lifelong}. However, this comes at the cost of significantly increasing the number of learnable parameters, making deployment more complex.
Our approach is fundamentally architecture-based but stands out by requiring only a small set of parameters (\emph{i.e.}, prompts) to be learned. This design allows for high efficiency in handling long task sequences without the need to store data, gradients, or models from previous tasks.

\subsection{Prompt Tuning}
PT \cite{karimi2021compacter,li2021prefix,gu2022ppt,wang2022multitask} is a lightweight approach for adapting PLMs \cite{li2025system,zhang2024mm,zhao2025chartcoder,zhao2025chartedit} to specific downstream tasks. This method involves optimizing a sequence of virtual tokens, known as soft prompts, which are appended to the input of PLMs, while keeping the PLMs themselves frozen. Recently, PT has been extended to the CL domain~\cite{qin2021lfpt5,DBLP:conf/acl/0007LM0H22,wang2022learning,wang2022dualprompt,wang2023rehearsal,DBLP:conf/acl/LiangWJQHH23,razdaibiedina2022progressive}. Among these methods, ProgPrompt~\cite{razdaibiedina2022progressive} stands out, achieving SOTA performance across various CTC benchmarks. It mitigates CF by learning distinct prompts for each task and facilitates FKT by progressively appending previously learned prompts as inputs when acquiring new tasks.

However, ProgPrompt has limitations. It overlooks the shared knowledge among tasks, and directly concatenating all learned prompts for FKT introduces redundant information. Furthermore, the growing prompt list increases computational overhead within the Transformer architecture. To address these issues, we propose InfoComp, which incorporates a S-Prompt to encode task-invariant knowledge shared across all tasks \cite{liu2023transcending,bai2024learning}, ensuring that only essential knowledge is utilized for FKT, thereby avoiding redundant information. In addition, the prompt list remains constant in length during CL, and our information theory framework promotes the learning of more informative prompts.

\subsection{Information Theory Approaches in Natural Language Processing}

Information theory, pioneered by Claude Shannon in the mid-20th century, provides a mathematical framework for quantifying information \cite{shannon1948mathematical,shannon1951prediction,cover1999elements}. It has played a crucial role in advancing various natural language processing tasks, including language modeling, machine translation, and information retrieval \cite{brown1992class,manning1999foundations,tishby2000information,steinborn2022information}. By measuring entropy and MI within language data, researchers have developed models that effectively capture linguistic structures and semantic relationships.

In text memorization, information-theoretic metrics help evaluate a model's ability to retain and reproduce textual information \cite{DBLP:conf/emnlp/WestHBC19,ju2021leveraging}. This involves quantifying the amount of preserved information during processing to prevent overfitting or underfitting, ensuring a balanced trade-off between memorization and generalization.

In multimodal translation, where textual and visual data are integrated, MI serves as a key criterion for aligning different modalities \cite{DBLP:conf/emnlp/JiZZHS22}. By maximizing MI, models can better capture cross-modal dependencies, improving translation accuracy and robustness across multimodal inputs.

For model pretraining, InfoBERT \cite{DBLP:conf/iclr/WangWCGJLL21} enhances BERT's robustness by applying the information bottleneck principle, which filters out task-irrelevant information to improve generalization and resilience against adversarial perturbations. 
Similarly, INFOXLM \cite{DBLP:conf/naacl/ChiDWYSWSMHZ21} employs MI-based optimization to refine cross-lingual language modeling, promoting semantic alignment across languages and enhancing zero-shot transfer capabilities. 
Furthermore, \cite{wei2024diff} introduces an evaluation metric based on the ``effective rank'' of model representations, drawing on insights from information theory and geometry to analyze and quantify how large language models eliminate redundant information throughout the training process. This approach offers a comprehensive assessment of model performance by capturing changes in representational efficiency before and after training.

For fine-tuning, \cite{mahabadi2021variational} introduces an information bottleneck approach to low-resource fine-tuning, retaining task-relevant information while minimizing redundancy to enhance efficiency and generalization. Moreover, \cite{DBLP:conf/acl/SorensenRRSRDKF22} and \cite{wu2023infoprompt} propose information-theoretic prompt engineering methods, leveraging MI to optimize prompt selection and improve language model conditioning.

Building on these methods, we leverage MI to quantify the information shared between two random variables, allowing us to compare prompts with other parameters as well as different encoded representations. This methodology facilitates the generation of more informative prompts, leading to improved task-specific and task-invariant knowledge representation for CTC.

\section{Preliminary}

\subsection{Problem Formulation}

Following ProgPrompt \cite{razdaibiedina2022progressive}, this work explores a CL scenario in which a PLM is required to handle a sequence of $n$ text classification tasks $(T_1, ..., T_n)$. Each task $T_k$ ($k = 1, \dots, n$) is composed of a set of \emph{i.i.d.} training examples $\{X^k_i, Y^k_i\}_{i=1}^{m_k}$, where $X^k_i$ denotes the $i$-th sequence of input text tokens, and $Y^k_i$ is the corresponding label drawn from a predefined set $\mathcal{Y}_k$. The PLM, parameterized by $\Theta$, is assumed to have access to task identity during both the training and inference phases.\footnote{We focus on the task-incremental setup, where task identity is generally assumed to be known during inference \cite{razdaibiedina2022progressive}.} Unlike many previous approaches, we do not assume access to data from previous tasks (\emph{i.e.}, Rehearsal-free). Therefore, the PLM is limited to using data from task $T_k$ only while training on that task. As a result, the learning objective across all tasks is defined as:
\begin{equation}
   \max_{\Theta} \; \sum_{k=1}^{n} \sum_{i=1}^{m_k} \log p_{\Theta}(Y^k_i| X^k_i)
\end{equation}

The simplest method for CL is finetuning, where the PLM sequentially minimizes the loss for each task $T_k$, $k \in \{1..n\}$, by updating all its parameters $\Theta$ (including the parameters of PLM and classifier):
\begin{equation}
\mathcal{L}_k(\Theta) =
 -\sum_{i=1}^{m_k}{\log p(Y^k_i | X^k_i, \Theta)}  
 \label{ft}
\end{equation}

While continual finetuning facilitates FKT to future tasks, it also leads to CF, where performance on previously learned tasks deteriorates after learning new ones, ultimately resulting in increased generalization loss \citep{kirkpatrick2017overcoming, de2019episodic, mccloskey1989catastrophic}.

\subsection{ProgPrompt}

To remedy this defect, ProgPrompt adopts a strategy of learning distinct prompts for each task to mitigate CF and facilitates FKT by progressively appending previous prompts (frozen) as inputs when learning a new task \cite{razdaibiedina2022progressive}. The training objective for the $k$-th task $T_k$ in Equation (\ref{ft}) is therefore modified as:
 \begin{equation}
\mathcal{L}_k(\theta_{P_k},\theta^{\text{head}}_k) =
 -\sum_{i=1}^{m_k}{\log p(Y^k_i | [P_k,\dots,P_1,X^k_i],\theta,\theta^{\text{head}}_k,\theta_{P_1},\dots,\theta_{P_k})} \text{,}
 \label{prog}
 \end{equation}
during training, the parameters $\theta$ of the PLM remain fixed, while only the current prompt-specific parameters $\theta_{P_k}$ and the current classifier parameters $\theta^{\text{head}}_k$ are updated for task $T_k$; they are frozen once the task $T_k$ is completed.

It becomes apparent from Equation (\ref{prog}) that ProgPrompt faces several limitations. First, it emphasizes task-specific knowledge while failing to exploit shared knowledge across different tasks. Second, its approach of concatenating all previously learned prompts to achieve FKT results in the inclusion of redundant information, which can reduce efficiency. Additionally, ProgPrompt continuously expands the list of prompts as new tasks are introduced. With the number of tasks ($N$) growing, the size of the prompt list increases linearly, at a rate of $\mathcal{O}(N)$. Given that transformer-based PLMs have a computational complexity of $\mathcal{O}(N^2)$~\cite{vaswani2017attention}, this leads to rising costs in both training and inference as the number of tasks accumulates. Consequently, the approach becomes less scalable over time.

\section{Method}

To address the aforementioned limitations, we introduce the InfoComp method, which combines a pair of complementary prompts with an information-theoretic framework. This approach is designed to efficiently guide PLMs in learning a sequence of downstream text classification tasks, enabling better performance in CL scenarios.

\subsection{Complementary Prompts}

The complementary learning systems theory \cite{mcclelland1995there,kumaran2016learning} in cognitive science suggests that humans rely on the interaction between the hippocampus and the neocortex to achieve CL. The hippocampus specializes in forming pattern-separated representations from specific, individual experiences, while the neocortex is responsible for extracting more general, transferable knowledge from a sequence of past experiences \cite{arani2021learning,pham2021dualnet,wang2022dualprompt}. This collaboration enables humans to effectively learn and adapt to new information while retaining previously acquired knowledge. 

\begin{figure*}[t]
\centering
  \includegraphics[width=1.0\linewidth]{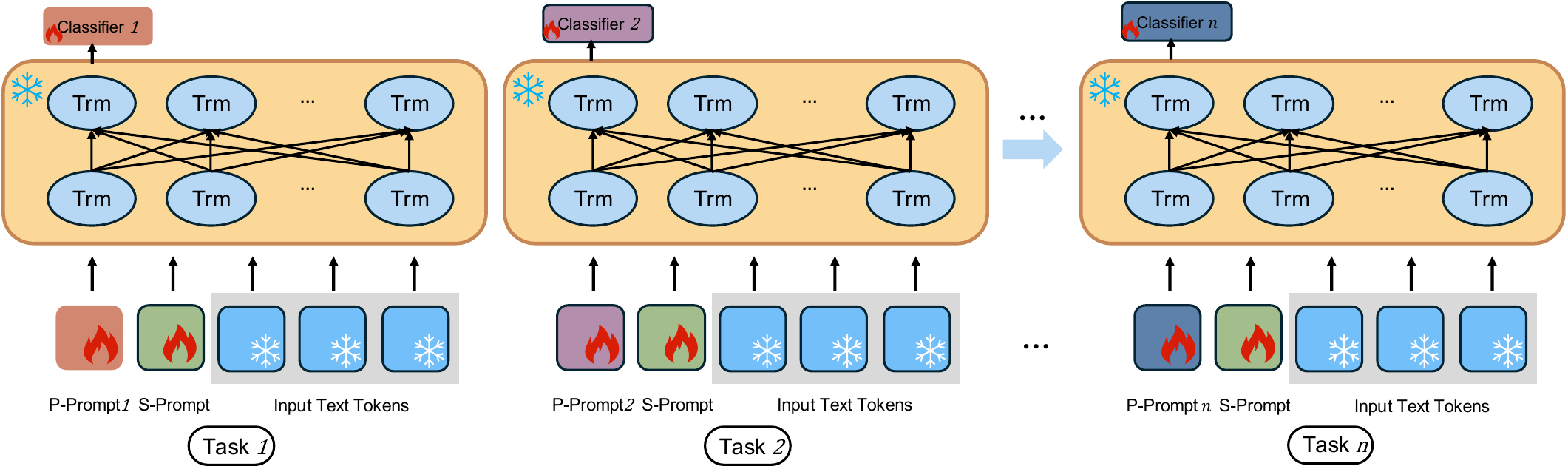} 
\caption{InfoComp learns two distinct prompt types: P-Prompt and S-Prompt. The P-Prompt encodes task-specific knowledge to reduce CF, while the S-Prompt captures task-invariant knowledge to enhance FKT. ``Trm" represents the Transformer encoder block within the PLM.}
\label{fig:main}
\end{figure*}

Building on this theory, InfoComp introduces a pair of complementary prompts: P-Prompt and S-Prompt, as illustrated in Figure \ref{fig:main}. For each task, a unique P-Prompt is learned to encode task-specific knowledge, helping to prevent interference between tasks and mitigate CF. Meanwhile, a shared S-Prompt is utilized across all tasks to capture task-invariant knowledge, facilitating FKT. The training objective for the $k$-th task $T_k$ is formulated as follows: 
 \begin{equation}
\mathcal{L}_k(\theta_{P_k},\theta_{S},\theta^{\text{head}}_k) =
 -\sum_{i=1}^{m_k}{\log p(Y^k_i | [P_k,S,X^k_i],\theta,\theta^{\text{head}}_k,\theta_{P_k},\theta_{S})}\text{,}
 \label{comp}
 \end{equation}
where $\theta_{P_k}$ represents the parameters associated with the task-specific P-Prompt $P_k$, while $\theta_{S}$ denotes the parameters linked to the S-Prompt $S$.

In contrast to ProgPrompt as defined in Equation (\ref{prog}), InfoComp in Equation (\ref{comp}) introduces an additional S-Prompt to capture task-invariant knowledge shared across all tasks. Rather than concatenating all previously learned P-Prompts for FKT, InfoComp leverages the generalized knowledge stored in the S-Prompt, effectively reducing redundant information. When training on a new task, only the corresponding P-Prompt and the shared S-Prompt are used, ensuring that the total prompt length remains fixed, even as the number of tasks ($N$) increases. This design keeps the model efficient and scalable, regardless of the number of tasks involved.

\subsection{Information-Theoretic Framework}

To further enhance the learning of more informative prompts, InfoComp introduces an information-theoretic framework, framing PT as a process of maximizing MI between prompts and other model parameters, as well as across different encoded representations. This framework leads to the design of two novel MI-based training objectives: one aimed at enhancing task-specific knowledge in P-Prompt, and another focused on preserving task-invariant knowledge in S-Prompt. These objectives are intended to balance the integration of task-relevant insights with the retention of generalized knowledge.

\paragraph{Enhancing Task-Specific Knowledge in P-Prompt}

Recent research has indicated that directly optimizing prompts alone is insufficient for encoding sufficient task-specific information \cite{wu2023infoprompt}. In contrast, task-specific classifiers are typically enriched with information from downstream tasks, and their parameters tend to adapt to task-specific knowledge more efficiently since they are closely aligned with the classification objectives. To explicitly enhance the task-specific information within the P-Prompt, we maximize the mutual information $I(\theta_{P_k};\theta^{\text{head}}_k|X^k_i)$ between the P-Prompt parameters $\theta_{P_k}$ and the classifier parameters $\theta^{\text{head}}_k$, thereby more effectively mitigating CF.

Given the simplicity of vector dot product calculations and their gradient-friendly nature, we directly maximize the inner product between the P-Prompt parameters $\theta_{P_k}$ and the classifier parameters $\theta^{\text{head}}_k$. This guides the model toward learning representations that are more closely related. By doing so, we effectively enhance the MI between these parameters. The process can be formalized as follows:
 \begin{equation}
\mathcal{L}_k^{\text{p-info}}(\theta_{P_k}, \theta^{\text{head}}_k) = -{\theta^{\text{head}}_k}^TW_1\theta_{P_k}^T\text{,}
 \label{p-info}
 \end{equation}
where $W_1$ is a trainable transformation matrix.

\paragraph{Preserving Task-Invariant Knowledge in S-Prompt}

Since the S-Prompt participates in the optimization of every task, it is inevitably affected by task-specific updates. To mitigate this influence and ensure the S-Prompt captures and retains as much task-agnostic, general knowledge as possible, we maximize the MI between the encoded representations of the same input using both the current and previous S-Prompt states. This strategy ensures better preservation of task-invariant knowledge, enhances FKT, and supports the continuity of shared knowledge across sequential tasks.

\begin{figure*}[t]
\centering
  \includegraphics[width=0.6\linewidth]{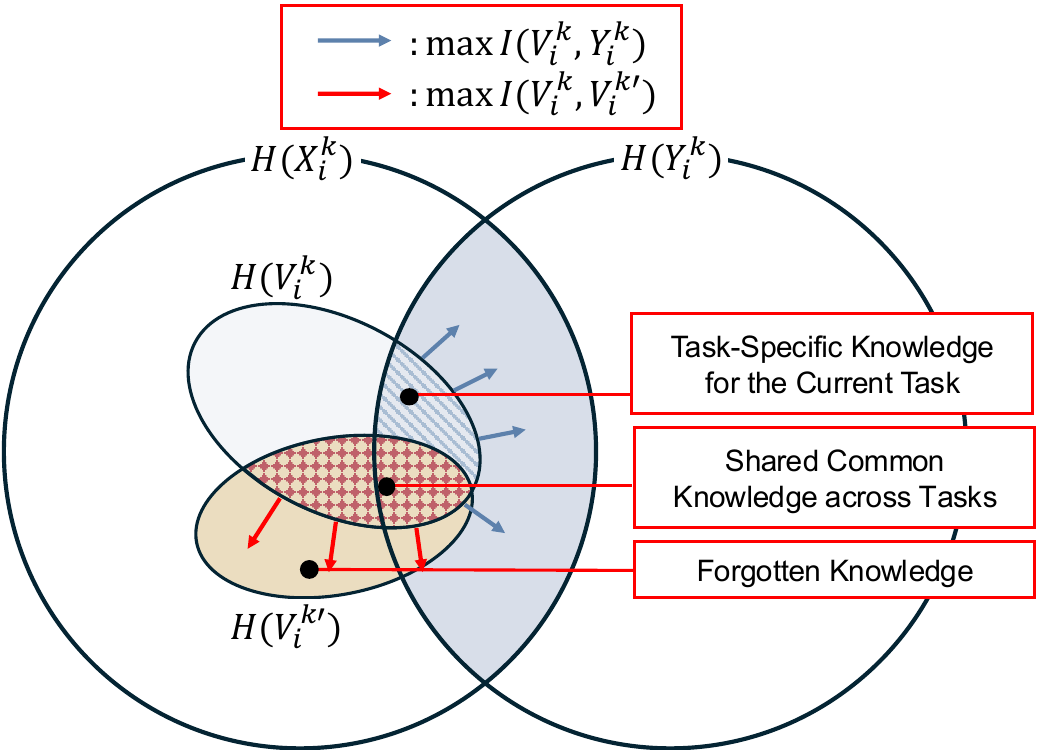} 
\caption{Alleviating the issue of forgetting shared knowledge across tasks by maximizing the MI $I(V^k_i, V^{k'}_i)$.}
\label{venn}
\end{figure*}

Let the PLM encoder be denoted as $F(\cdot)$. Given an input pair $(X^k_i, Y^k_i)$ from the current task $T_k$, we define $V^k_i = F(X^k_i, S)$ as the representation generated using the current S-Prompt $S$, and $V^{k'}_i = F(X^k_i, S')$ as the representation produced by the frozen S-Prompt $S'$, which was the S-Prompt at the end of the previous task optimization. 
Consider a Markov chain $V^{k'}_i \leftarrow X^k_i \rightarrow V^{k}_i  \rightarrow Y^k_i$. From an information-theoretic perspective, optimizing for each task involves maximizing the MI $I(V^k_i, Y^k_i)$. As illustrated in Figure \ref{venn}, this process increases the amount of task-specific knowledge (represented by the blue arrow), while reducing the shared common knowledge across tasks. To address the issue of forgetting shared knowledge within the S-Prompt, we need to introduce an explicit constraint by maximizing the MI $I(V^k_i, V^{k'}_i)$, following the direction of the red arrow.

\begin{figure*}[t]
\centering
  \includegraphics[width=1.0\linewidth]{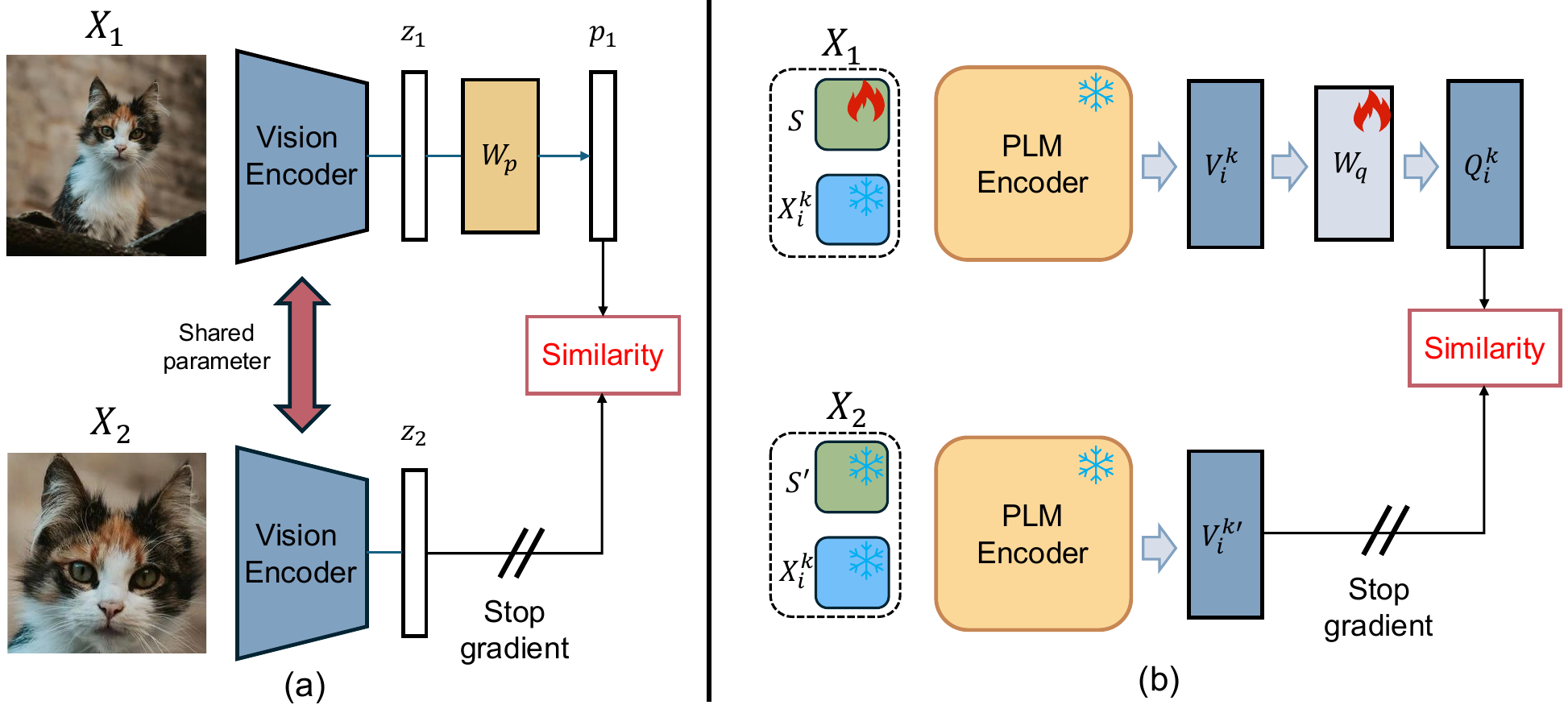} 
\caption{(a) SimSiam for the positive sample pair. (b) SimSiam for the prompt-augmented positive sample pair.}
\label{simsiam}
\end{figure*}

Inspired by SimSiam \cite{chen2021exploring}, as shown in Figure \ref{simsiam} (a), it is a positive-only contrastive learning method that captures invariant features from positive sample pairs. Its training objective is formalized as follows:
 \begin{equation}
\mathcal{L}^{\text{SimSiam}} =-(W_pz_1)^Tz_2 = -p_1^Tz_2\text{,}
 \end{equation}
where $W_p$ is a trainable matrix designed to capture a shared feature embedding space between different views of positive samples (For example, as shown in Figure \ref{simsiam} (a), it depicts different views of the same cat.).

By analogy, we can view $X_1 = (X^k_i, S)$ and $X_2 = (X^k_i, S')$ as a pair of positive sample views of the same input $X^k_i$ under different S-Prompt conditions from different tasks, as shown in Figure \ref{simsiam} (b). To increase the MI $I(V^k_i, V^{k'}_i)$, we can adopt the SimSiam training objective, bringing this pair of positive samples closer in the embedding space to enhance the interdependence between $V^k_i$ and $V^{k'}_i$. This can be formalized as follows:
 \begin{equation}
\mathcal{L}^{\text{s-info}}_k(\theta_S) = -(W_qV^k_i)^TV^{k'}_i= - {Q^k_i}^TV^{k'}_i \text{,}
 \label{s-info}
 \end{equation}
where $W_q$ is a trainable matrix designed to capture the shared subspace between tasks. The parameters of the current S-Prompt, denoted as $\theta_S$, are optimized to generate an embedding space that aligns with the S-Prompt from the previous task. This guides the current S-Prompt to expand the shared subspace with prior tasks, thereby enhancing the retention of previously learned general knowledge.

\subsection{Overall Training Objective}

Finally, the overall training objective function of our proposed InfoComp method is formulated as follows:
 \begin{equation}
\mathcal{L}^{\text{overall}}_k = \mathcal{L}_k + \lambda_1 \mathcal{L}^{\text{p-info}}_k + \lambda_2 \mathcal{L}^{\text{s-info}}_k \text{,}
 \end{equation}
where $\lambda_1$ and $\lambda_2$ are hyperparameters that balance the importance of the loss terms.

\section{Experimental Settings}

\subsection{Datasets}

\begin{table}[t]
    \centering
\caption{The description of the $15$ datasets utilized in our CTC experiments. ``Acc." stands for accuracy. The first five tasks are part of the standard CTC benchmark, while the remaining tasks are included in our long-sequence experiments.}
\label{tab:datasets_explained}
\begin{tabular}{lcccc}
\toprule
 \textbf{Name} & \textbf{Alias} & \textbf{Source} & \textbf{\#.Class} &  \textbf{Metric} \\
 \midrule
 1. AG News & ag & Standard  & $4$  & Acc.\\
 2. Amazon reviews & amazon & Standard  & $5$  & Acc.\\
 3. DBpedia &dbpedia& Standard  & $14$  & Acc.\\
  4. Yelp reviews &yelp& Standard  & $5$  & Acc.\\
 5. Yahoo Answers &yahoo& Standard  & $10$  & Acc.\\
 6. MNLI & mnli& GLUE & $3$  & Acc.\\
 7. QQP &qqp & GLUE & $2$  & F1 \\
 8. RTE &rte& GLUE & $2$  & Acc. \\
 9. SST2 & sst2& GLUE & $2$  & Acc. \\
 10. WiC & wic& SuperGLUE & $2$  & Acc. \\
 11. CB & cb& SuperGLUE & $3$  & Acc. \\
 12. COPA & copa& SuperGLUE & $2$  & Acc. \\
13. MultiRC & multirc& SuperGLUE & $2$  & Acc. \\
 14. BoolQ & boolq& SuperGLUE & $2$  & Acc. \\
 15. IMDB movie& imdb & Other & $2$  & Acc. \\
 \bottomrule
\end{tabular}
\end{table}

\paragraph{Standard CTC Benchmark}

Building on the methodology outlined in ProgPrompt \cite{razdaibiedina2022progressive}, we initially assess our proposed approach using the widely recognized CTC benchmark, which comprises five text classification datasets introduced by \cite{zhang2015character}: AG News, Amazon reviews, DBpedia, Yelp reviews, and Yahoo Answers. In line with earlier CTC studies such as IDBR \cite{huang2021continual} and MBPA++ \cite{de2019episodic}, we conduct evaluations across four distinct task orderings of these five datasets. The training and testing sets are identical to those used in IDBR \cite{huang2021continual} and MBPA++ \cite{de2019episodic}, comprising $115,000$ training and $7,600$ testing instances per task. Following the approach in \cite{huang2021continual}, for each task we randomly withhold $500$ samples per class from the training set for validation, applying early stopping based on validation accuracy.

\paragraph{Long-sequence CTC Benchmark}

A more realistic CTC scenario involves longer task sequences with a larger number of tasks. To address this, we also assess InfoComp's performance on a more challenging benchmark comprising $15$ text classification tasks \cite{razdaibiedina2022progressive}. This benchmark includes the five datasets from the standard CTC benchmark, along with four tasks from the GLUE benchmark (MNLI, QQP, RTE, SST2) \cite{wang2018glue}, five tasks from the SuperGLUE benchmark \cite{wang2019superglue} (WiC, CB, COPA, MultiRC, BoolQ), and the IMDB movie reviews dataset \cite{maas2011learning}. In line with ProgPrompt \cite{razdaibiedina2022progressive}, we evaluate performance across varying dataset sizes by creating three versions of each dataset, containing $20$, $200$, and $1000$ training examples per class, and report test results for each case. Consistent with the approach in \cite{huang2021continual}, for each task we randomly set aside $500$ samples per class from the training data for validation, using early stopping based on validation accuracy.

\begin{table}[t]
\centering

\caption{Seven distinct task sequence orders were used for the CTC experiments. Orders $1$ through $4$ correspond to the standard CTC benchmark, while Orders $5$ through $7$ represent long-sequence settings involving $15$ tasks.}
\begin{tabular}{cc}
\toprule
\textbf{Order}  & \textbf{Task Sequence} \\
\midrule
$1$  & ag $\arr$ yelp $\arr$ amazon $\arr$ yahoo $\arr$ db \\
$2$  & yelp $\arr$ yahoo $\arr$ amazon $\arr$ db $\arr$ ag \\
$3$  & db $\arr$ yahoo $\arr$ ag $\arr$ amazon $\arr$ yelp \\
$4$ & yelp $\arr$ ag $\arr$ db $\arr$ amazon $\arr$ yahoo \\
\midrule
$5$ & \makecell{mnli $\arr$ cb $\arr$ wic $\arr$ copa $\arr$ qqp $\arr$ boolq $\arr$ rte $\arr$ imdb $\arr$ \\ yelp $\arr$ amazon $\arr$ sst2 $\arr$ dbpedia $\arr$ ag $\arr$ multirc $\arr$ yahoo} \\
$6$ & \makecell{multirc $\arr$ boolq $\arr$ wic $\arr$ mnli $\arr$ cb $\arr$ copa $\arr$ qqp $\arr$ rte $\arr$ \\ imdb $\arr$ sst2 $\arr$ dbpedia $\arr$ ag $\arr$ yelp $\arr$ amazon $\arr$ yahoo} \\
$7$  & \makecell{yelp $\arr$ amazon $\arr$ mnli $\arr$ cb $\arr$ copa $\arr$ qqp $\arr$ rte $\arr$ imdb $\arr$  \\ sst2 $\arr$ dbpedia $\arr$ ag $\arr$ yahoo $\arr$ multirc $\arr$ boolq $\arr$ wic} \\
\bottomrule
\end{tabular}
\label{tab:seq}
\end{table}

Table~\ref{tab:datasets_explained} provides a detailed overview of the $15$ datasets employed in our CTC experiments. These datasets are drawn from the standard CTC benchmark \citep{zhang2015character}, the GLUE \citep{wang2018glue} and SuperGLUE \citep{wang2019superglue} benchmarks, with the addition of the IMDB movie reviews dataset. The task orders used in our CTC experiments are presented in Table \ref{tab:seq}.

\subsection{Baselines}

We compare InfoComp against eight baseline methods\footnote{Consistent with \cite{huang2021continual,razdaibiedina2022progressive}, in all replay methods, $1$\% of samples per class are stored (with a minimum of 1 sample for smaller datasets).}:

\begin{itemize}
    \item \textbf{Finetune} \cite{wang2020efficient, de2019episodic, huang2021continual}: all model parameters are trained sequentially on a series of tasks, without applying any regularization constraints or replaying samples from earlier tasks.

    \item \textbf{Experience Replay}: the entire model is fine-tuned using a memory buffer, with samples from previous tasks replayed during the learning of new tasks to mitigate forgetting. 

    \item \textbf{A-GEM} \cite{chaudhry2018efficient}: store examples from previous tasks and limit the gradients used for updating the model on new tasks by referencing the retrieved examples.

    \item \textbf{MBPA++} \cite{de2019episodic}: enhance BERT with an episodic memory that stores all encountered examples. Replay occurs during training, while local adaptation is applied during testing.
    
    \item \textbf{IDBR} \cite{huang2021continual}: this BERT-specific method continuously trains the entire model by incorporating experience replay along with a regularization loss, which separates sentence representations into task-specific and task-generic components.

    \item \textbf{Per-task Prompts} \cite{lester2021power}: a separate soft prompt is trained for each task, while the original model's parameters remain unchanged. This method effectively avoids CF by ensuring that task-specific parameters are preserved as new tasks are introduced. However, it lacks FKT, as shared knowledge across tasks is not leveraged.
    
    \item \textbf{Prompt Tuning} \cite{lester2021power, qin2021lfpt5}: only a shared soft prompt is trained sequentially across all tasks, with the original model parameters kept fixed throughout.

    \item \textbf{InfoCL} \cite{song2023infocl}: a novel replay-based CTC approach that leverages fast-slow and current-past contrastive learning to enhance the representation of diverse knowledge.

    \item \textbf{Q-Tuning} \cite{guo2024q}: incorporates a prompt queue alongside an adaptive knowledge aggregation low-rank matrix, which is optimized to assess the significance of stored prompts and improve FKT.

    \item \textbf{SLM} \cite{bohaoscalable}: offers a model-agnostic framework for the scalable acquisition of knowledge and skills. It integrates vector space retrieval into the language model and features two key components: joint adaptive re-parameterization and dynamic retrieval of task-specific knowledge.

    \item \textbf{ProgPrompt} \cite{razdaibiedina2022progressive}: it learns unique prompts for each new task and reuses them during inference, effectively reducing CF. Additionally, it enables FKT by using previously learned, frozen prompts as inputs when training on a new task.

\end{itemize}

\subsection{Evaluation Metric}

Following ProgPrompt \cite{razdaibiedina2022progressive}, we use average accuracy on the test set as our evaluation metric, representing the overall average accuracy across all tasks after training on the final task. Additionally, for certain task sequences, we include detailed task-wise accuracy plots. These plots capture the average accuracy of all tasks completed up to that point, measured after each task is trained. To assess the statistical significance of our improvements, we conduct a paired t-test at a significance level of $0.05$ \cite{koehn2004statistical}.

\subsection{Implementation Details}

For consistency with prior works, including IDBR \cite{huang2021continual}, MBPA++ \cite{de2019episodic}, and ProgPrompt \cite{razdaibiedina2022progressive}, we adopt the pre-trained bert-base-uncased model as the PLM backbone. Our implementation relies on PyTorch \cite{paszke2019pytorch} along with the HuggingFace Transformers library \cite{wolf2019huggingface}. 

For the standard CTC benchmark, we use the official datasets from \cite{zhang2015character}\footnote{\url{http://goo.gl/JyCnZq}}, consistent with \cite{de2019episodic,zhang2015character}. Additionally, we obtain data for GLUE tasks \cite{wang2018glue}, SuperGLUE tasks \cite{wang2019superglue}, and the IMDB movie reviews dataset \cite{maas2011learning} using the HuggingFace Datasets library\footnote{\url{https://github.com/huggingface/datasets}}, leveraging these datasets for long-sequence CTC experiments. In line with previous studies \cite{rao2019continual,de2019episodic}, we repurpose each dataset's validation set as the test set (where test data is unavailable) and reserve 500 samples from the training set to create the validation set.

\begin{table}[t]
    \centering
    \begin{tabular}{lcccccc} 
    \toprule
      & & \multicolumn{4}{c}{\textbf{Order}} & \\
      \textbf{Baseline} & ER & \textbf{1} & \textbf{2}  & \textbf{3}  & \textbf{4} & \textbf{Avg.} \\
      \midrule
      Finetune & & 14.8 & 27.8 & 26.7 & 4.5  & 18.4 \\
      Experience Replay & \checkmark  & 67.2 & 64.7 & 64.7 & 44.6 & 57.8 \\
      A-GEM  & \checkmark & 70.6 & 65.9 & 67.5 & 63.6 & 66.9 \\
      MBPA++ & \checkmark  & 70.8 & 70.9 & 70.2 & 70.7 & 70.6 \\
      IDBR  & \checkmark & 75.9 & 76.2 & 76.4 & 76.7 & 76.3 \\

  InfoCL$^{*}$ & \checkmark& 76.5& 76.8& 70.0 & 76.3& 74.9\\
      
      ProgPrompt$^{\diamondsuit}$ & & 78.0 & 77.7 & 77.9 & 77.9  & 77.9 \\

    ProgPrompt$^{\diamondsuit,*}$ & & 78.0 & 77.8 & 77.8 & 77.8  & 77.9\\

Q-Tuning$^{\diamondsuit}$ & & 78.5 &78.3& 78.3 &78.4 &78.4\\

  SLM$^{\diamondsuit}$ & & \textcolor{blue}{\textbf{79.2}} &\textcolor{blue}{\textbf{78.8}} &\textcolor{blue}{\textbf{79.0}} &\textcolor{blue}{\textbf{79.2}} & \textcolor{blue}{\textbf{79.1}}\\

       \midrule
    \textbf{InfoComp (Ours)}$^{\diamondsuit}$ & & $\textcolor{deepred}{\textbf{79.6}}^{\dagger}$ & $\textcolor{deepred}{\textbf{80.1}}^{\dagger}$ & $\textcolor{deepred}{\textbf{79.9}}^{\dagger}$& $\textcolor{deepred}{\textbf{80.2}}^{\dagger}$ & $\textcolor{deepred}{\textbf{80.0}}^{\dagger}$ \\
    \textbf{Improve} & & $\Uparrow$\textbf{0.4}&  $\Uparrow$\textbf{1.3}&$\Uparrow$\textbf{0.9}&$\Uparrow$\textbf{1.0}&$\Uparrow$\textbf{0.9}\\

      \bottomrule
    \end{tabular}
\caption{Summary of results on the standard CTC benchmark. The highest result is highlighted in \textcolor{deepred}{\textbf{red}}, while the second-highest result is shown in \textcolor{blue}{\textbf{blue}}. A $\dagger$ marker indicates a statistically significant result with a $p$-value of $<0.05$ in comparison to ProgPrompt \cite{razdaibiedina2022progressive} and SLM \cite{bohaoscalable}. The column ``ER" specifies whether experience replay is required by each baseline. 
Baselines labeled with $^{\diamondsuit}$ maintain a frozen PLM, while the remaining methods train the entire model. Those marked with $*$ are derived from our re-implementation using their open-source code, whereas the other baseline results are directly cited from ProgPrompt \cite{razdaibiedina2022progressive} or their respective original papers.}
    \label{tab:standard}
\end{table}

We employ the Adam optimizer \cite{kingma2014adam} for all experiments, setting a batch size of $8$ and a learning rate of $1e$-$4$. Prompts are trained for between $40$ and $300$ epochs, based on the dataset size. Specifically, for the standard CTC benchmark, we set the epoch count to $40$. For the long-sequence CTC benchmark, the epoch count is set to $300$, $150$, and $40$ when each class contains $20$, $200$, and $1000$ samples, respectively. The final prompts are selected based on the best validation set scores obtained from the prompt checkpoints. Following \cite{lester2021power}, prompts are initialized with randomly sampled tokens. In all experiments, we configure the P-Prompt length, G-Prompt length, and the loss coefficients $\lambda_1$ and $\lambda_2$ to be $35$, $5$, $0.05$, and $0.1$, respectively\footnote{The prompt length refers to the sequence length of tokens in a prompt.}. All of our results are the average of $3$ runs. For all other baselines, we use hyperparameters outlined in their original papers.

\section{Experimental Results}

\begin{figure*}[tbp]
\centering
  \includegraphics[width=0.75\linewidth]{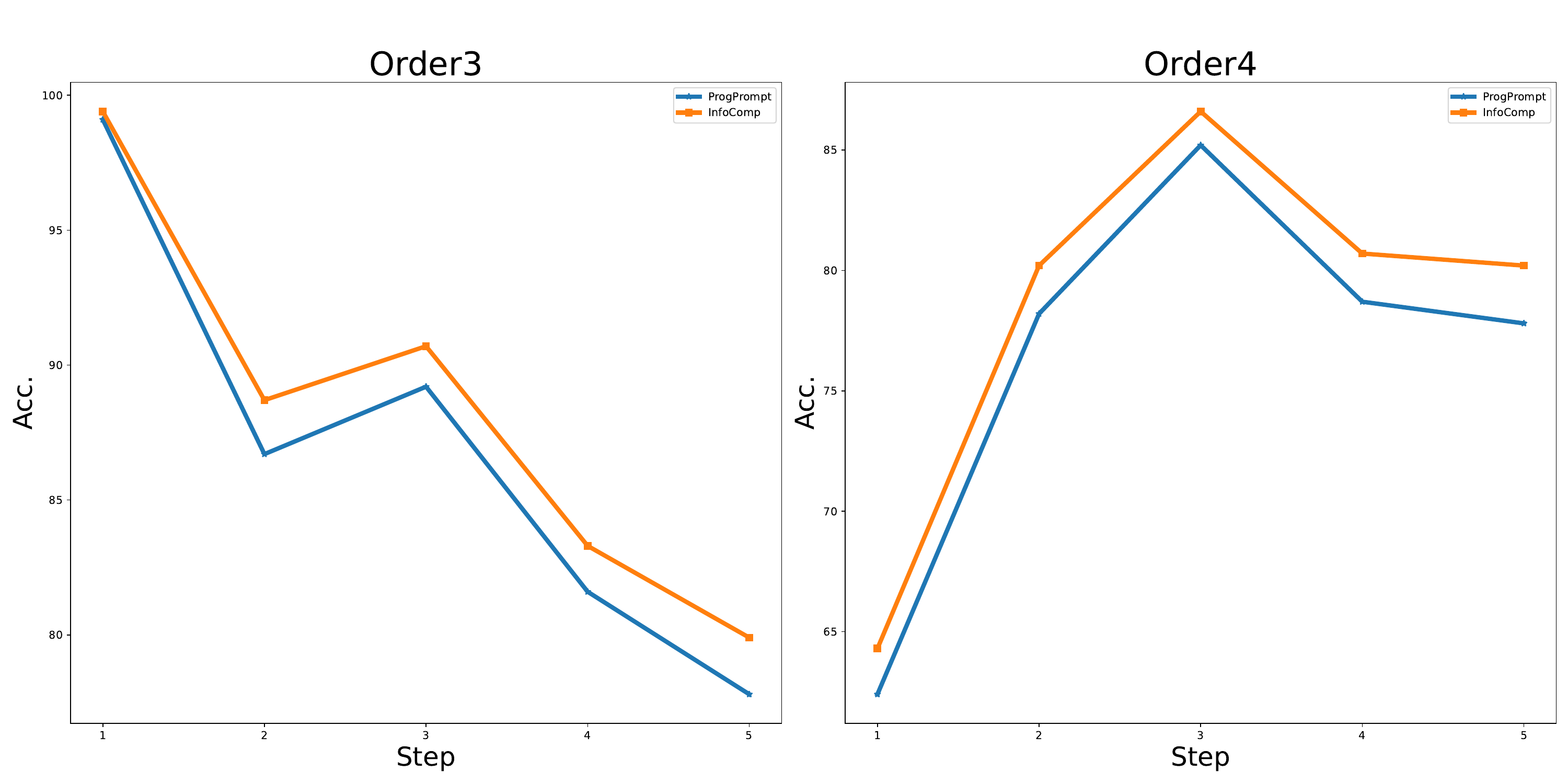}
\caption{Comparison of task-wise accuracy for the task sequences Order3 and Order4. Our InfoComp consistently outperforms ProgPrompt \cite{razdaibiedina2022progressive}, across all task-wise evaluations.}
    \label{fig:step}
\end{figure*}

\subsection{Standard CTC Benchmark}

\subsubsection{Main Results} 

Table \ref{tab:standard} presents a comparison of InfoComp's performance against four different task sequences and existing CTC methods, including the previous SOTA approaches, ProgPrompt \citep{razdaibiedina2022progressive} and SLM \cite{bohaoscalable}. Overall, InfoComp enhances CTC performance, achieving scores of $79.6$, $80.1$, $79.9$, and $80.2$ across the four sequences, with an average score of $80.0$. This represents an improvement of $1.6$, $2.3$, $2.0$, and $2.3$ over ProgPrompt, and $0.4$, $1.3$, $0.9$, and $1.0$ over SLM, respectively. Furthermore, unlike traditional CTC methods such as IDBR \cite{huang2021continual}, MBPA++ \cite{de2019episodic}, and InfoCL \cite{song2023infocl}, InfoComp does not require the storage of data from previous tasks for future replay. These findings highlight the effectiveness of InfoComp.

\subsubsection{Task-Wise Comparisons}

Figure \ref{fig:step} illustrates the task-wise comparison results that comprehensively assess the effectiveness of our InfoComp method in CTC scenarios. The findings reveal that InfoComp consistently surpasses ProgPrompt \cite{razdaibiedina2022progressive}, across all task-wise comparisons. This enhanced performance is particularly pronounced in the experimental setups for task sequences Order3 and Order4. These results emphasize the robustness of InfoComp in sustaining high performance and adaptability while tackling new tasks throughout the learning process. The marked improvement over ProgPrompt highlights InfoComp's potential to set a new benchmark for CTC tasks, particularly in complex, real-world data environments.

\subsection{Long-Sequence CTC Benchmark}

\begin{table}[t]
\fontsize{7.5}{10}\selectfont
  \begin{center}
  \setlength{\tabcolsep}{3.0pt}
    \begin{tabular}{lccc|ccc|ccc|ccc} 
    \toprule
      \textbf{Baseline} $\downarrow$ & \multicolumn{3}{c}{\textbf{Order 5}}  & \multicolumn{3}{c}{\textbf{Order 6}}  & \multicolumn{3}{c}{\textbf{Order 7}} & \multicolumn{3}{c}{\textbf{Avg.}} \\
       \textbf{Num. samples} $\rightarrow$ & 20 & 200 & 1000  & 20 & 200 & 1000  & 20 & 200 & 1000 & 20 & 200 & 1000 \\
      \toprule
      Finetune & 29.9 & 43.4 & 40.9 & 30.5 & 42.0 & 42.5 & 33.6 & 41.9 & 41.8 & 31.3 & 42.4 & 41.7 \\
     Prompt Tuning$^{\diamondsuit}$ & -- & -- & -- & -- & -- & -- & -- & -- &  -- & 47.6 & 57.2 & 59.5 \\ 
      Experience Replay & 34.9 & 46.3 & 51.0 & 39.3 & 48.1 & 51.5  & 34.9 & 46.5 & 46.3 & 36.4 & 47.0 & 49.6 \\
       Per-task Prompts$^{\diamondsuit}$ & 50.6 & 62.4 & 67.2 & 50.6 & 62.4 & 67.2 & 50.6 & 62.4 & 67.2 & 50.6 & 62.4 & 67.2 \\
      IDBR       & 39.7 & 48.4 & 52.3 & 37.9 & 46.6 & 54.1 & 32.9 & 48.8 & 50.1 & 36.8 & 47.9 & 52.2 \\
       ProgPrompt$^{\diamondsuit}$ & \textcolor{blue}{\textbf{55.3}} & \textcolor{blue}{\textbf{67.9}} & \textcolor{blue}{\textbf{68.9}} & 53.3 & 65.8 & \textcolor{blue}{\textbf{70.0}} & 51.9 & \textcolor{blue}{\textbf{66.9}} & \textcolor{blue}{\textbf{69.0}} & 53.5 & \textcolor{blue}{\textbf{66.9}} & \textcolor{blue}{\textbf{69.3}} \\
    ProgPrompt$^{\diamondsuit,*}$ & 51.4 & 66.8 & 68.4 & \textcolor{blue}{\textbf{54.2}} & \textcolor{blue}{\textbf{66.9}} & 67.5 & \textcolor{blue}{\textbf{57.1}} & 66.4 & 68.7 & \textcolor{blue}{\textbf{54.2}} & 66.7 & 68.2 \\
 
      \midrule
    \textbf{InfoComp (Ours)}$^{\diamondsuit}$ &  $\textcolor{deepred}{\textbf{56.4}}^{\dagger}$ & $\textcolor{deepred}{\textbf{69.6}}^{\dagger}$ & $\textcolor{deepred}{\textbf{70.4}}^{\dagger}$& $\textcolor{deepred}{\textbf{57.7}}^{\dagger}$ & $\textcolor{deepred}{\textbf{69.2}}^{\dagger}$ & $\textcolor{deepred}{\textbf{72.4}}^{\dagger}$& $\textcolor{deepred}{\textbf{59.7}}^{\dagger}$& $\textcolor{deepred}{\textbf{69.9}}^{\dagger}$& $\textcolor{deepred}{\textbf{70.8}}^{\dagger}$& $\textcolor{deepred}{\textbf{57.9}}^{\dagger}$& $\textcolor{deepred}{\textbf{69.6}}^{\dagger}$& $\textcolor{deepred}{\textbf{71.2}}^{\dagger}$\\
    \textbf{Improve} &  $\Uparrow$\textbf{1.1}&  $\Uparrow$\textbf{1.7}&$\Uparrow$\textbf{1.5}&$\Uparrow$\textbf{3.5}&$\Uparrow$\textbf{2.3} &$\Uparrow$\textbf{2.4}&$\Uparrow$\textbf{2.6}&$\Uparrow$\textbf{3.0}&$\Uparrow$\textbf{1.8}&$\Uparrow$\textbf{3.7}&$\Uparrow$\textbf{2.7}&$\Uparrow$\textbf{1.9}\\
      \bottomrule
    \end{tabular}
\caption{Overview of results on the long-sequence CTC benchmark. The top result is highlighted in \textcolor{deepred}{\textbf{red}}, while the second-best is marked in \textcolor{blue}{\textbf{blue}}. $\dagger$ denotes statistical significance with a $p$-value less than $0.05$ compared to ProgPrompt \cite{razdaibiedina2022progressive}. Results are reported across different data limits: $20$, $200$, and $1000$ samples per class. Baselines marked with $^{\diamondsuit}$ indicate only soft prompts are trained while the PLM remains frozen, whereas other methods involve training the entire model. Results marked with $*$ come from our re-implementation, while other baseline results are directly sourced from ProgPrompt \cite{razdaibiedina2022progressive}.}
  \label{tab:long}
  \end{center}
\end{table}

Table~\ref{tab:long} presents a comparison of various common CTC approaches, including the SOTA method ProgPrompt \cite{razdaibiedina2022progressive}, within the context of sequence continual learning involving $15$ tasks. We provide averaged results across three task orders ($5$, $6$, and $7$) and include the complete non-averaged results for each order. To examine the impact of limited data settings, we conduct training using different dataset sizes of $20$, $200$, and $1000$ samples per class. Our method, InfoComp, consistently outperforms all competing approaches across these varying data constraints, achieving average scores of $57.9$, $69.6$, and $71.2$ for the few-shot scenarios of $20$, $200$, and $1000$ samples per class, respectively. This results in improvements of $3.7$, $2.7$, and $1.9$ over the previous SOTA method, ProgPrompt.

\subsection{Ablation Study}

\begin{table}[tbp]
  \centering
  \caption{The ablation study of our InfoComp under the task sequences Order6 and Order7. In comparison to our InfoComp method, all ablation variants show a significant decline in CTC performance, confirming the necessity of each component for collaboratively addressing CTC. \textbf{Bold} denotes the best results.}
  \resizebox{1.0\linewidth}{!}{
    \begin{tabular}{lccccccc}
    \toprule
    \multirow{2}[4]{*}{Variants} &  \multicolumn{3}{c}{Order6} & &\multicolumn{3}{c}{{Order7}}\\
\cmidrule{2-4} \cmidrule{6-8}
& 20 & 200  & 1000 & & 20 & 200  & 1000 \\
    \midrule
InfoComp w/o P-Prompt &48.2 & 57.9 & 60.3& & 49.1 & 58.2 & 61.1 \\

InfoComp w/o S-Prompt  & 51.2 & 63.0 & 68.7 & & 52.4 & 64.1 & 68.1  \\
    \midrule
InfoComp w/o $\mathcal{L}_k^{\text{p-info}}$\&$\mathcal{L}_k^{\text{s-info}}$  & 54.4 & 65.7 & 70.2 & & 57.3 & 66.1 & 68.8  \\

InfoComp w/o $\mathcal{L}_k^{\text{p-info}}$ in Equation (\ref{p-info})    &  56.5 & 67.6 & 71.3 & & 59.0 & 68.6 & 70.0  \\

InfoComp w/o $\mathcal{L}^{\text{s-info}}_k$ in Equation (\ref{s-info})  & 55.6 & 67.0 & 70.7 & &  58.2 & 67.4 & 69.5  \\

  \midrule

   \textbf{InfoComp (Ours)}  & \textbf{57.7} & \textbf{69.2} & \textbf{72.4} & & \textbf{59.7} & \textbf{69.9} & \textbf{70.8}  \\
    \bottomrule
    \end{tabular}
    }
  \label{tab:ablation}
\end{table}

This subsection assesses the effectiveness of the individual components of our InfoComp method in the task sequences Order6 and Order7 through ablation studies, as illustrated in Table~\ref{tab:ablation}. When we remove P-Prompt and S-Prompt from InfoComp, it indicates that only a shared soft prompt is trained sequentially across all tasks, or that a separate soft prompt is trained for each individual task. The first approach neglects task-specific knowledge, rendering it susceptible to CF for prior tasks, while the second approach overlooks shared knowledge among tasks, which can impede FKT to some extent. Both strategies result in a significant drop in performance compared to the complete InfoComp method, highlighting that effectively addressing CTC tasks requires simultaneously capturing both task-specific and task-invariant knowledge.

When we remove $\mathcal{L}_k^{\text{p-info}}$ and $\mathcal{L}_k^{\text{s-info}}$ from the InfoComp method, it indicates that the information-theoretic framework is not utilized to support the learning of prompts during training, which in turn affects the generation of more informative prompts. The individual removal of $\mathcal{L}_k^{\text{p-info}}$ compromises the enhancement of task-specific knowledge acquisition in P-Prompt, while the exclusion of $\mathcal{L}_k^{\text{s-info}}$ undermines the preservation of task-invariant knowledge in S-Prompt. As a result, these variations lead to a significant decline in CTC performance, underscoring the importance of each component in our InfoComp method for effectively addressing CTC challenges.

\section{Conclusion}

In this paper, we introduced InfoComp, a novel approach to CTC that effectively balances the learning of task-specific and task-invariant knowledge. InfoComp utilizes two distinct prompt spaces, P-Prompt and S-Prompt, enabling models to sequentially acquire new classification tasks while minimizing CF. Furthermore, by leveraging an information-theoretic framework, our designed loss functions significantly enhance the accumulation of task-specific knowledge in P-Prompt and improve the retention of task-invariant knowledge in S-Prompt. This reduces CF and fosters better FKT. Extensive experiments across various CTC benchmarks, including standard setups and more challenging scenarios with longer task sequences, demonstrated that InfoComp consistently outperforms existing SOTA methods, underscoring its effectiveness in addressing the challenges of continual learning. Future work could investigate the application of our model-agnostic InfoComp method with additional PLMs beyond BERT, allowing for further validation of its efficacy across a broader range of CTC benchmarks.

 \bibliographystyle{elsarticle-num} 
 \bibliography{ref.bib}

\end{document}